\documentclass{article}

\usepackage{PRIMEarxiv}

\usepackage[utf8]{inputenc} 
\usepackage[T1]{fontenc}    
\usepackage{hyperref}       
\usepackage{url}            
\usepackage{booktabs}       
\usepackage{amsfonts}       
\usepackage{nicefrac}       
\usepackage{microtype}      
\usepackage{lipsum}
\usepackage{fancyhdr}       
\usepackage{graphicx}       
\graphicspath{{media/}}     
\usepackage{amsmath, xparse}
\usepackage{xcolor}
\usepackage{color}
\usepackage{colortbl}

\title{Rethinking the Multi-view Stereo from the Perspective of Rendering-based Augmentation
}

\author{
  Chenjie Cao, Xinlin Ren, Xiangyang Xue, Yanwei Fu \\
  School of Data Science, Fudan University \\
  \texttt{\{20110980001,20110240015,xyxue,yanweifu\}@fudan.edu.cn} \\
}

\begin{document}
\maketitle

\begin{abstract}
GigaMVS presents several challenges to existing Multi-View Stereo (MVS) algorithms for its large scale, complex occlusions, and gigapixel images. 
To address these problems, we first apply one of the state-of-the-art learning-based MVS methods, --MVSFormer, to overcome intractable scenarios such as textureless and reflections regions suffered by traditional PatchMatch methods, but it fails in a few large scenes' reconstructions. 
Moreover, traditional PatchMatch algorithms such as ACMMP, OpenMVS, and RealityCapture are leveraged to further improve the completeness in large scenes.
Furthermore, to unify both advantages of deep learning methods and the traditional PatchMatch, we propose to render depth and color images to further fine-tune the MVSFormer model.
Notably, we find that the MVS method could produce much better predictions through rendered images due to the coincident illumination, which we believe is significant for the MVS community. 
Thus, MVSFormer is capable of generalizing to large-scale scenes and complementarily solves the textureless reconstruction problem.
Finally, we have assembled all point clouds mentioned above \textit{except ones from RealityCapture} and ranked Top-1 on the competitive GigaReconstruction.
\end{abstract}



\section{Introduction}
\label{sec:intro}

\begin{figure}[h!]
  \centering
  \includegraphics[width=0.95\linewidth]{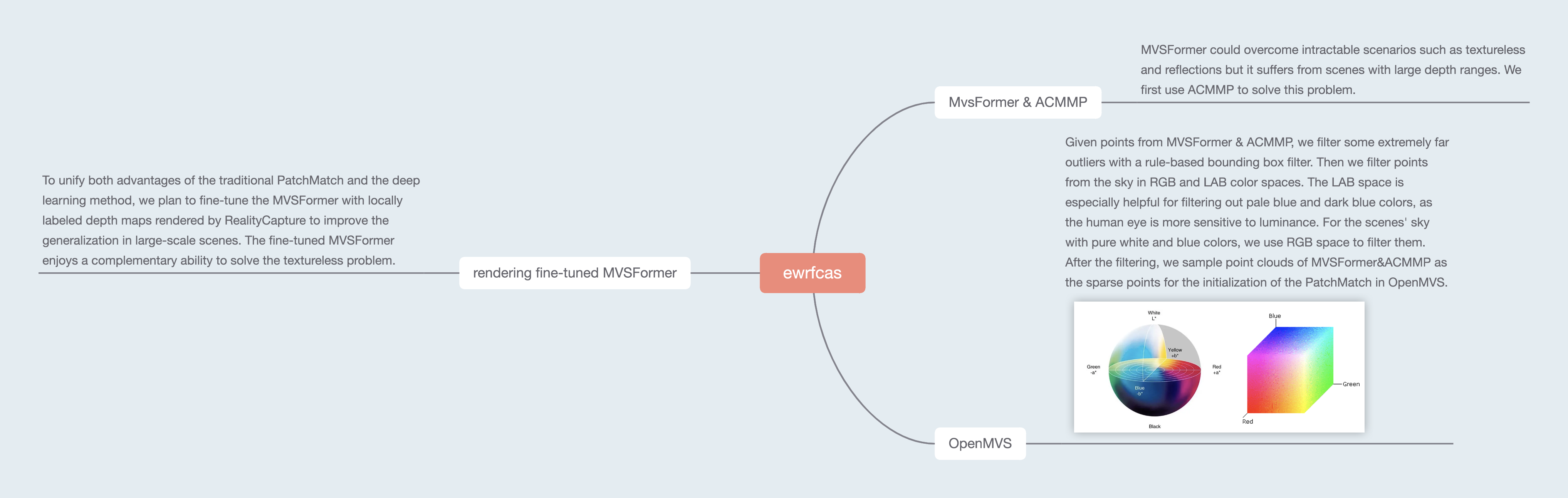} 
  \caption{The overview of our submission. Our results comprise four parts, which include MVSFormer~\cite{cao2023mvsformer} \& ACMMP~\cite{xu2022multi}, OpenMVS~\cite{openmvs}, and rendering fine-tuned MVSFormer.}
  \label{fig:overview}
\end{figure}

This technical report presents our competition plan in the GigMVS reconstruction~\footnote{https://gigavision.cn}. Besides, we also introduce our technical contributions to this competition as well as the important finding for the community.

The final point cloud submission of our team consists of MVSFormer~\cite{cao2023mvsformer} (Sec.~\ref{sec:mvsformer}) \& ACMMP~\cite{xu2022multi} (Sec.~\ref{sec:acmmp}), OpenMVS~\cite{openmvs}\footnote{https://github.com/cdcseacave/openMVS} (Sec.~\ref{sec:openmvs}), 
and fine-tuned MVSFormer based on 
the Capture Reality's\footnote{https://www.capturingreality.com} rendering with BlenderProc~\cite{denninger2019blenderproc} (Sec.~\ref{sec:render}). Our method is over-viewed in Fig.~\ref{fig:overview}. For these Multi-View Stereo (MVS) methods, our MVSFormer is a learning-based model while others are based on the classical PatchMatch~\cite{zheng2014patchmatch} algorithm. \textit{Note that point clouds from CaptureReality did not join in our final ensemble submission.}
More importantly, we find that \textit{rendered images enjoy much better MVS predictions due to the coincident illustration} (Sec.~\ref{sec:finding}).

\section{Key Models: MVSFormer\&ACMMP}

\subsection{MVSFormer}
\label{sec:mvsformer}
MVSFormer, a state-of-the-art multi-view stereo (MVS) method~\cite{cao2023mvsformer}, employs several advanced techniques, including pre-trained Vision Transformer (ViT)~\cite{chu2021twins} models, multi-scale training, and temperature-based depth prediction to enable robust point cloud reconstruction. 
MVSFormer could overcome intractable scenarios such as textureless and reflections of high-resolution images with robust depth and confidence maps, which lead to high-quality point clouds.
MVSFormer outperforms other state-of-the-art MVS methods on the DTU dataset~\cite{aanaes2016large} and the immediate set of Tanks-and-Temple dataset~\cite{knapitsch2017tanks} with appropriate depth ranges.
Hence we select MVSFormer as our learning-based method to tackle the GigaReconstruction.

\begin{figure}
\includegraphics[width=0.95\linewidth]{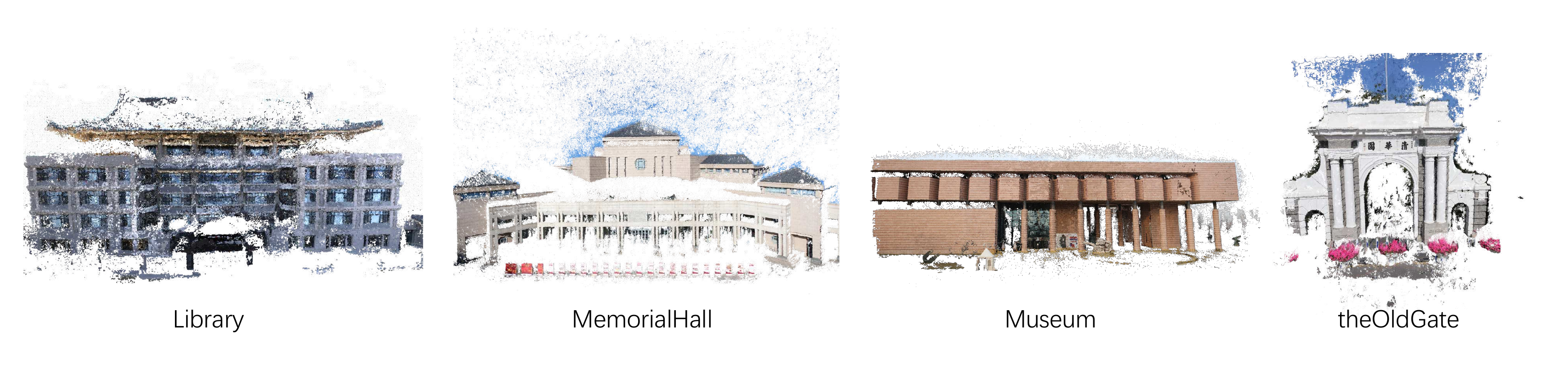} 
  \centering
  \caption{Qualitative point cloud results of MVSFormer.}
  \label{fig:mvsformer-quali}

\end{figure}

Specifically, MVSFormer is used to solve five scenes, which include `Museum', `ScienceSquare', `theOldGate', `Library', and `MemorialHall'. MVSFormer can achieve good results in these scenes because their min-max depth ranges are not very large. Detailed settings are shown in Tab.~\ref{tab:mvsformer_setting}, and qualitative results are shown in Fig.~\ref{fig:mvsformer-quali}. Since improving the recall score is more difficult in this task, we empirically chose hyper-parameters for more complete point clouds. Such a strategy is also adopted for other methods in our solution.

\begin{table}
\small 
\caption{The hyper-parameter setting of MVSFormer. `Conf' means the confidence threshold for depth filtering.
\label{tab:mvsformer_setting}}
\centering
\begin{tabular}{cccc}
\toprule 
Scene & Resolution & Conf & View Number\tabularnewline
\midrule
Museum & 2048 & 0.5 & 20\tabularnewline
ScienceSquare & 2048 & 0.5 & 20\tabularnewline
theOldGate & 2048 & 0.3 & 20\tabularnewline
Library & 4096 & 0.5 & 10\tabularnewline
MemorialHall & 4096 & 0.5 & 10\tabularnewline
\bottomrule
\end{tabular}
\end{table}

\begin{figure}
\includegraphics[width=0.95\linewidth]{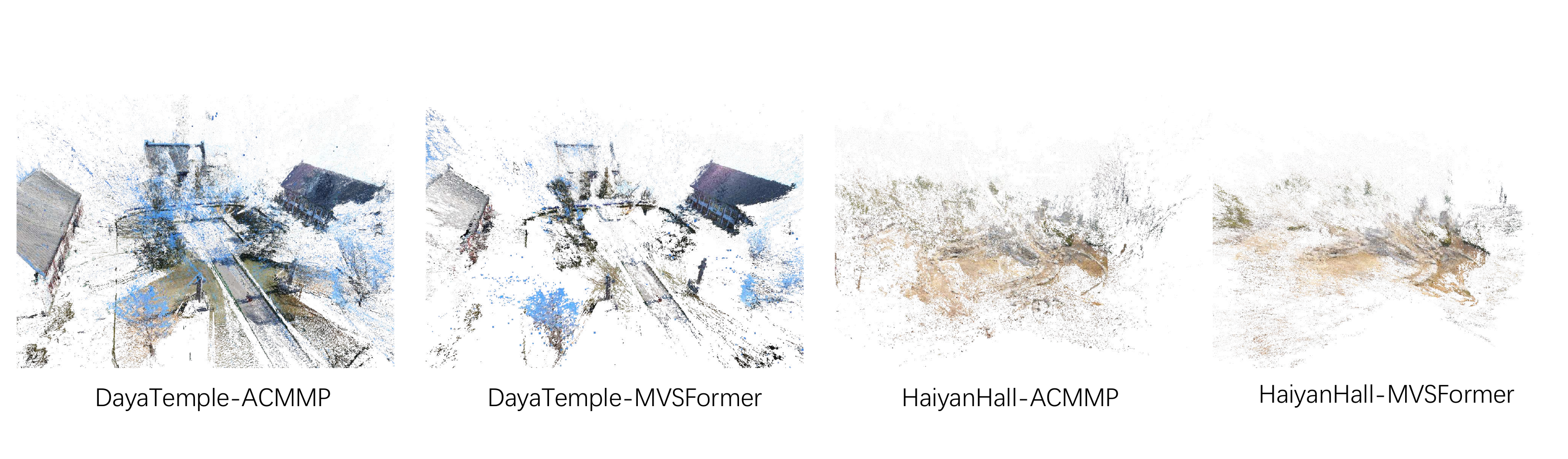} 
  \centering
  \caption{Comparison between MVSFormer and ACMMP.}
  \label{fig:acmmp-vs-mvsformer}
\end{figure}

\subsection{ACMMP}
\label{sec:acmmp}
The large depth range of GigaMVS~\cite{zhang2021gigamvs} is not suitable for CascadeMVS~\cite{gu2020cascade} and other learning-based methods to build cost volumes with limited depth intervals. 
We use a traditional PatchMach Stereo method --ACMMP~\cite{xu2022multi} for the rest three scenes with large depth variations (`DayaTemple', `PeonyGarden', `HaiyanHall'). In particular, the series of ACMM (ACMH, ACMM, ACMP)~\cite{Xu2019ACMM,Xu2020ACMP,xu2022multi} have achieved good results in large scene reconstruction on ETH3D~\cite{schops2017multi}. 
As one of the typical PatchMatch-based methods, ACMMP adopts a more robust and efficient checkerboard strategy for view selection based on COLMAP~\cite{schonberger2016pixelwise} and Gipuma~\cite{Galliani_2015_ICCV}.
ACMMP also leverages multi-scale geometric consistency and plane priors to enhance the performance of depth/normal predictions. Following our code document\footnote{https://github.com/maybeLx/Gigmvs-ewrfcas}, it is easy to run the ACMMP. Moreover, it achieves better reconstruction 
in some scenes compared with the learning-based MVSFormer as shown in Fig.~\ref{fig:acmmp-vs-mvsformer}. But the reconstruction of ACMMP is relatively slower.


\section{Technical Implementations by OpenMVS}
\label{sec:openmvs}
OpenMVS~\cite{openmvs} is a well-known open-source MVS library. 
As OpenMVS focuses on mesh reconstruction, it integrates many optimizations for reconstructing high-precision point clouds.
Therefore, we utilize the module of dense point-cloud reconstruction in OpenMVS. Specifically, we sample point clouds of MVSFormer\&ACMMP as the sparse points for the initialization of the PatchMatch in OpenMVS.

\subsection{Pre-processing for Sparse Point Clouds}
\begin{figure}
\includegraphics[width=0.95\linewidth]{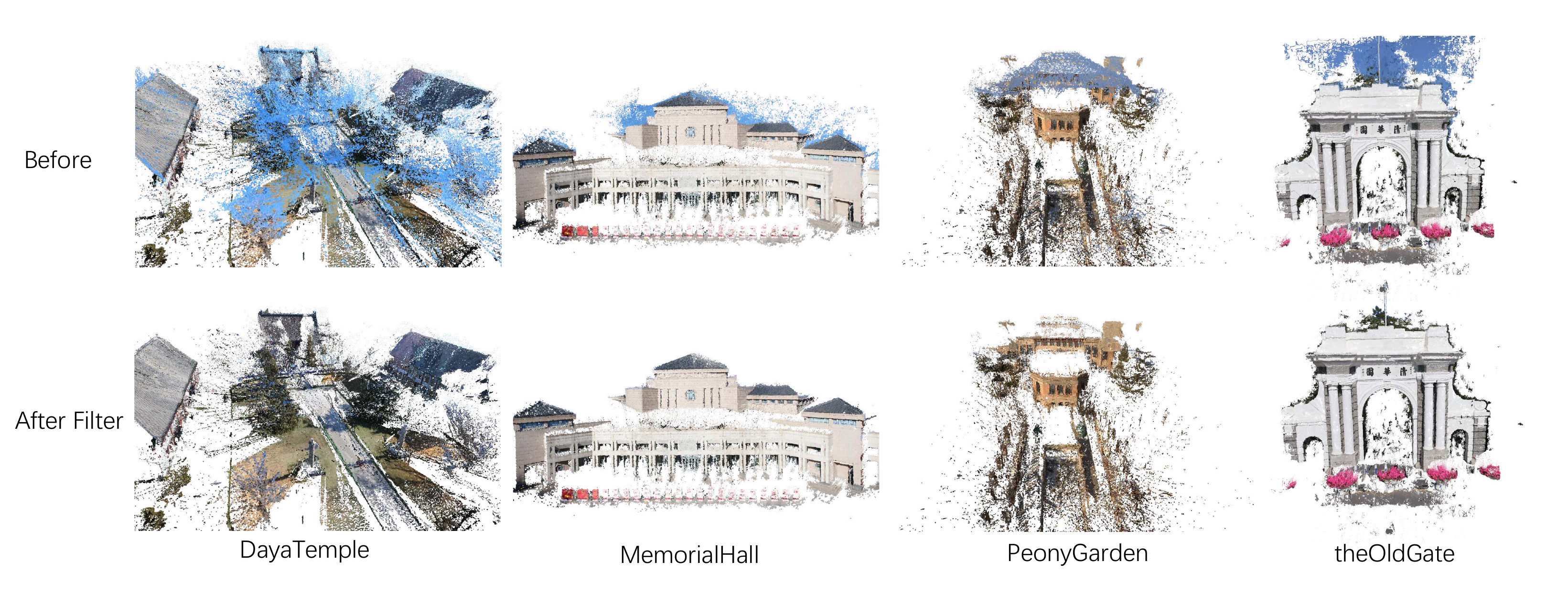} 
  \centering
  \caption{Visualization of filtered point cloud and original point cloud .}
  \label{fig:acmmp-filter}
\end{figure}

Before the sparse point sampling, we first filter some extremely far outliers with rule-based methods. Then, we filter points from the sky in RGB and LAB\footnote{https://en.wikipedia.org/wiki/CIELAB\_color\_space} color spaces as in Fig.~\ref{fig:acmmp-filter}.

\begin{figure}
  \centering
  \includegraphics[width=0.95\linewidth]{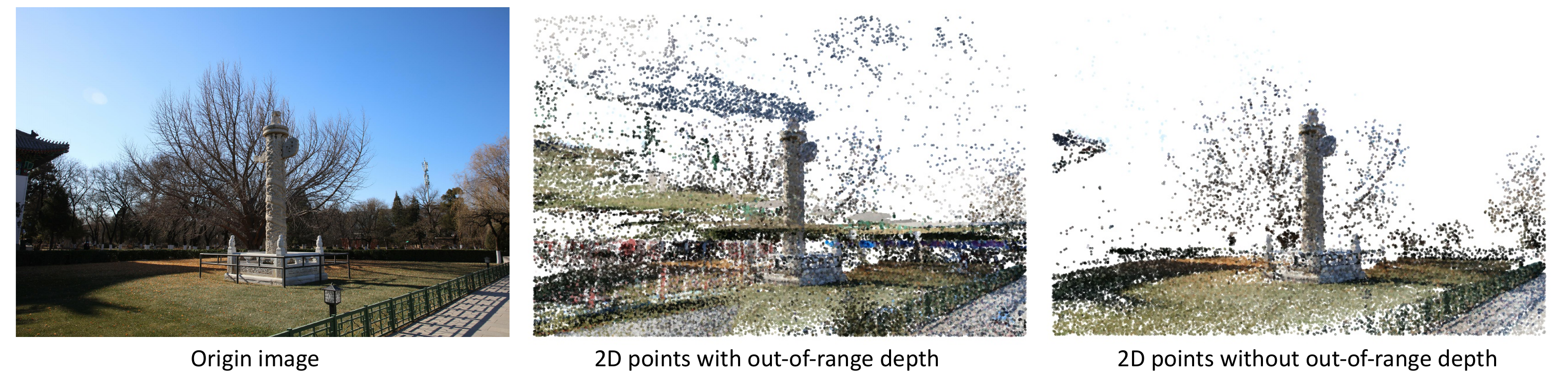} 
  \caption{Visualization of sampled sparse points for a view of `DayaTemple'.}
  \label{fig:sparse_points}
\end{figure}

To prepare the sparse points for OpenMVS, we should distribute valid points for each view. Furthermore, each 3D point must be confirmed with a unique id as COLMAP~\cite{schonberger2016pixelwise}. Given 3D point $\hat{p}_i$ from MVSFormer\&ACMMP, intrinsic matrix $\mathbf{K}$, extrinsic rotation $\mathbf{R}$, translation $t$, we can achieve 2D point $p_i$ as

\begin{equation}
\label{eq:3d-to-2d}
\tilde{p}_i=\mathbf{K}(\mathbf{R}\hat{p}_i+t),\quad
d_i=\tilde{p}_i[2],\quad
p_i=\tilde{p}_i[0:2]/d_i,
\end{equation}

where $d_i$ can be seen as the depth of point $i$ in this view. In principle, we should delete all wrapped 2D points with $d_i<0$. But in practice, we find that remaining these `out-of-range' points helps to reconstruct more complete dense point clouds. Thus we retain these points as shown in Fig.~\ref{fig:sparse_points} except the `PeonyGarden'. To balance the computation, we sample 500k 3D points uniformly for each scene.

\begin{figure}
  \centering
  \includegraphics[width=0.95\linewidth]{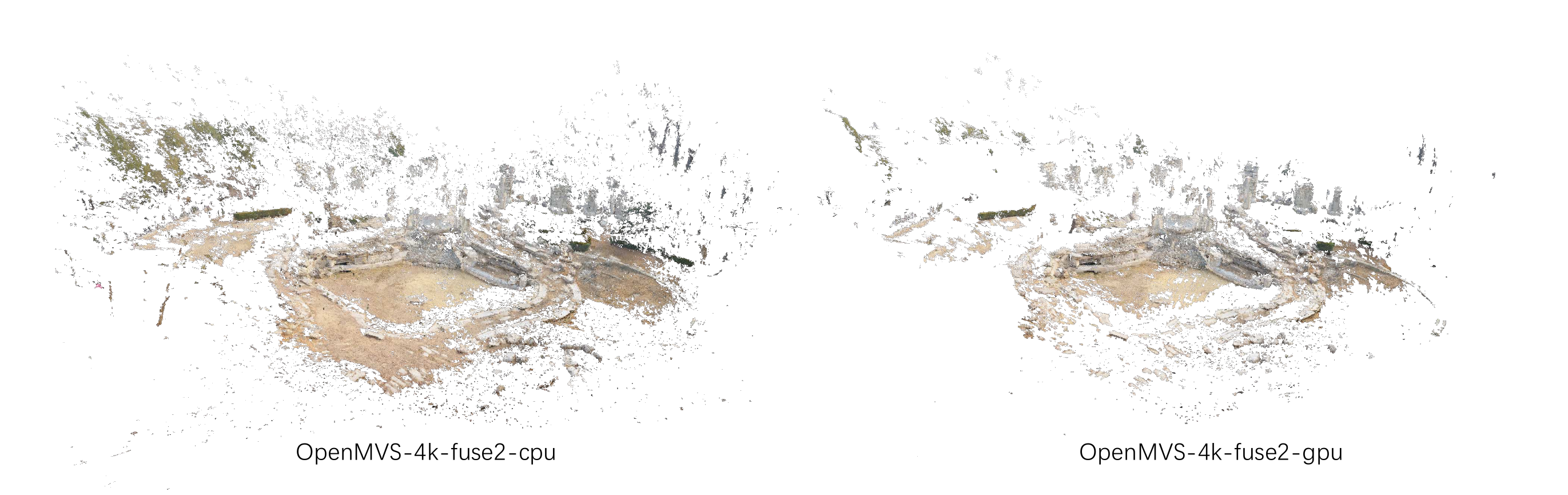} 
  \caption{Visualization of point clouds reconstructed by OpenMVS in different deivce environments }
  \label{fig:openmvs_cpu_gpu}
\end{figure}

\subsection{Settings and Environments of OpenMVS}
We adjust a few hyper-parameters for the dense reconstruction of OpenMVS as shown in Tab.~\ref{tab:setting_of_openmvs}. And we ensemble these four groups of OpenMVS results. In general, different resolutions enjoy better diversity for the ensemble. Moreover, we find that the implementation of OpenMVS-CPU and OpenMVS-GPU is different in PatchMatch, which leads to disparate point clouds as shown in Fig.~\ref{fig:openmvs_cpu_gpu}. Thus we also ensemble the results reconstructed with different device environments.

\begin{table}
\small 
\caption{The hyper-parameter setting of OpenMVS. Other settings are all set as default.
\label{tab:setting_of_openmvs}}
\centering
\begin{tabular}{ccc}
\toprule 
Resolutions & Fusion views & Env\tabularnewline
\midrule
2560(2k) & 2 & CPU\tabularnewline
4096(4k) & 2 & CPU\tabularnewline
4096(4k) & 2 & GPU\tabularnewline
8192(8k) & 2 & GPU\tabularnewline
\bottomrule
\end{tabular}
\end{table}

\section{Data Augmentation by RealityCapture and BlenderProc}
\label{sec:capturereality}
\begin{figure}
  \centering
  \includegraphics[width=0.95\linewidth]{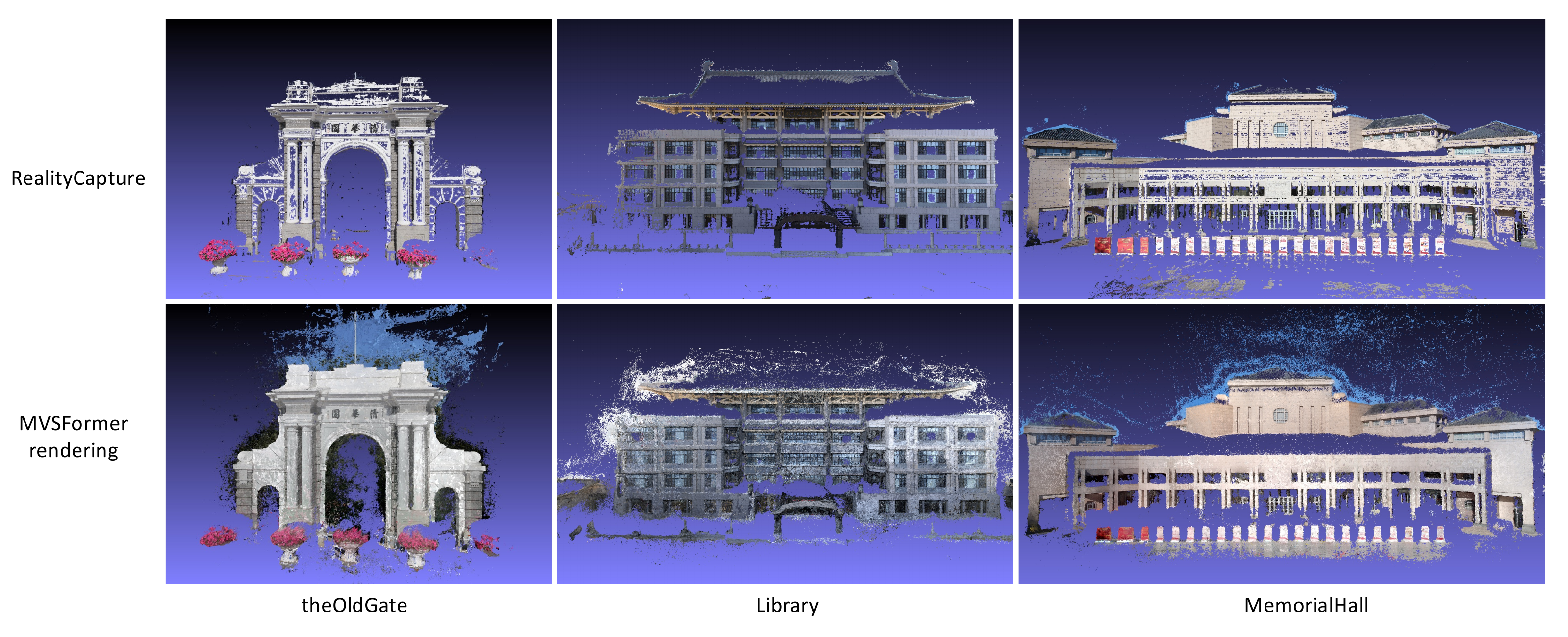} 
  \caption{Visualization of point cloud in textureless regions reconstructed by RealityCapture and our rendering fine-tuned MVSFormer.}
  \label{fig:capture_reality}
\end{figure}

RealityCapture is based on traditional PatchMatch manners with better quality in reconstructing textured mesh compared with OpenMVS. However, this method is also discouraged by the poor performance of textureless regions as shown in Fig.~\ref{fig:capture_reality}. We use RealityCapture with the 1.0 version, while all settings are set as default during the mesh reconstruction.

\subsection{Point Cloud Registration (Alignment).}
To render images in Sec.~\ref{sec:render}, 
we follow the process of ~\cite{knapitsch2017tanks} and align the RealityCapture's point clouds to the OpenMVS ones (4k-gpu). 
Formally, we first use camera poses to initialize the similarity transformation, \textit{i.e.}, sim(3) refinement.
Then we use the Iterative Closest Point (ICP)~\cite{besl1992method} to further refine the registration of the dense point clouds.
For the sim(3) refinement, given camera position $p_i, q_i$ from GigaMVS and RealityCapture respectively, 
we need to get a similarity transformation $\textbf{T}\in\mathbb{R}^{4\times4}$ as
\begin{equation}
\label{eq:transformation_T}
\textbf{T}=
\begin{bmatrix} 
c\textbf{R} & \textbf{t} \\
0 & 1 \end{bmatrix},
\end{equation} 
which is used to roughly warp RealityCapture's cameras $q_i$ to GigaMVS ones $p_i$. And the similarity transformation could be optimized by the following objective:
\begin{equation}
\label{eq:sim_3_refine}
E(\textbf{T})=\sum_{i=0}^{N} \Vert p_i-\textbf{T}q_i\Vert^2,
\end{equation}
where the points are represented in homogeneous coordinates; $N$ indicates the number of all valid camera poses. 
However, during the reconstruction of RealityCapture, it may drop some images, resulting in unmatched camera poses compared with GigaMVS ones. This phenomenon potentially causes problems in the alignment process. To address this issue, we remove the corresponding GigaMVS camera poses that are omitted by RealityCapture from $N$ before the alignment process.
For the ICP, we further finely align the dense point clouds from RealityCapture and GigaMVS twice as in~\cite{knapitsch2017tanks}.

\subsection{Rendering with BlenderProc}
\label{sec:render}

\begin{figure}
  \centering
  \includegraphics[width=0.95\linewidth]{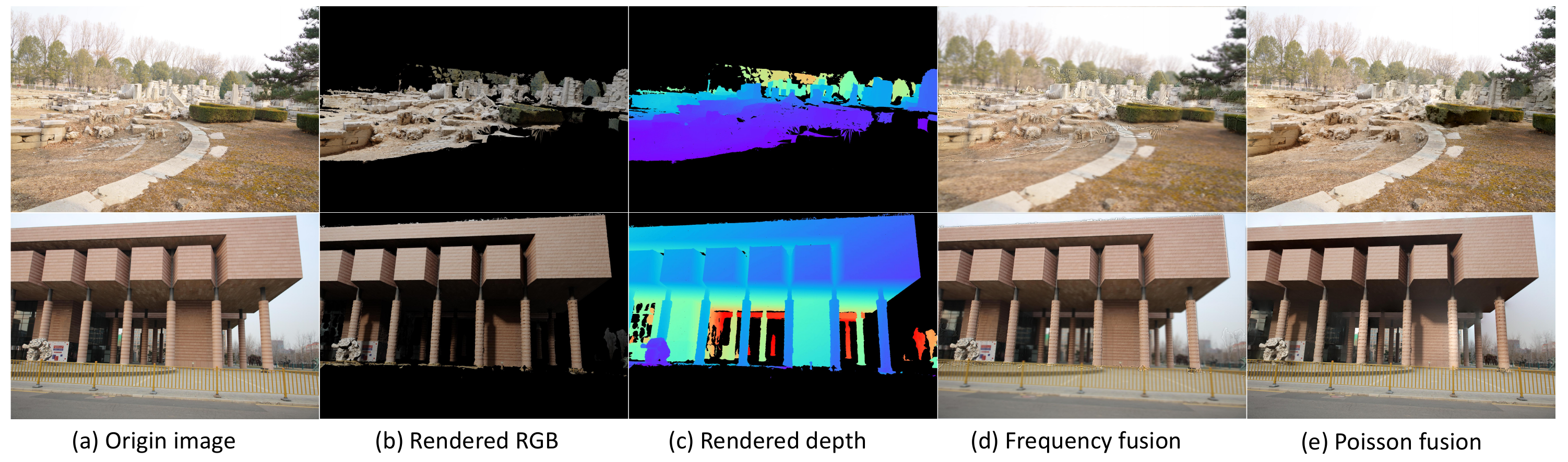} 
  \caption{Rendering of RealityCapture's mesh based on BlenderProc. We also compare the fusion manners with (d) frequency fusion in BlendedMVS~\cite{yao2020blendedmvs} and (e) our Poisson-based one.}
  \label{fig:rendering}
\end{figure}

To unify both advantages of traditional PatchMatch and deep learning methods, we follow the idea of BlendedMVS~\cite{yao2020blendedmvs} to render depth and RGB color images with RealityCapture's mesh and camera poses provided by GigaMVS. Then we plan to fine-tune the MVSFormer with locally labeled depth maps to improve the generalization of MVSFormer in large-scale scenes. Besides, the learning-based MVSFormer enjoys a complementary ability to solve the textureless problem. 
Formally, we use BlenderProc~\cite{denninger2019blenderproc} to render color images and depth as shown in Fig.~\ref{fig:rendering}. Since the mesh reconstruction is non-watertight, we filter mesh faces $f_i$ based on the face area $A_i$, the length of the longest side $S_{max}$, and the shortest side $S_{min}$ as follows
\begin{equation}
\label{eq:mesh_filter}
f_i
\left\{  
\begin{array}{rcl}
A_i<0.05, & \\
S_{max}\leq 0.25  , &\\
\frac{S_{max}}{S_{min}}\leq 7. &   
\end{array}
\right.
\end{equation}
Besides, we find that the lighting position is significant for reality during the rendering. Note that we abandon two scenes, `DayaTemple' and `PenoyGarden'. Because there are too many missing regions of these two scenes' meshes for training. But we still test our rendering-based model on `PenoyGarden', which enjoys acceptable results.

Because of the difficult large-scale scenes and limited views, the mesh quality of RealityCpature is far from the one of BlendedMVS. So directly combining real images and rendered images in the frequency fields with 2D Fast Fourier Transformation (FFT) as in~\cite{yao2020blendedmvs} causes obvious artifacts.
Different from BlenderedMVS, we use Poisson blending~\cite{perez2003poisson} to fuse rendered images and original images as in Fig.~\ref{fig:rendering}(e). And these images and rendered sparse depth are used to fine-tune the MVSFormer.
All rendered images and depth are in 1/4 of the original resolution. During the multi-scale training in MVSFormer, we further reduce the maximum resolution to 1088$\times$1664. Besides, we randomly apply the original RGB images, normal clone-based Poisson fusion, or monochrome transfer-based Poisson fusion as the training images to improve the training diversity. Moreover, fine-tuning with both BlendedMVS and rendered GigaMVS simultaneously benefits the final performance.
More fine-tuning details are in our released codes.

After fine-tuning, MVSFormer can produce high-quality depth on the training images. The fine-tuned MVSFormer also improves the performance on textureless regions as shown in Fig.~\ref{fig:capture_reality}. During the inference, we empirically find that rendered RGB images enjoy better depth predictions compared with original ones.

\section{Images based on Lambertian and Non-Lambertian Materials in MVS}
\label{sec:finding}

\begin{figure}[h!]
  \centering
  \includegraphics[width=0.9\linewidth]{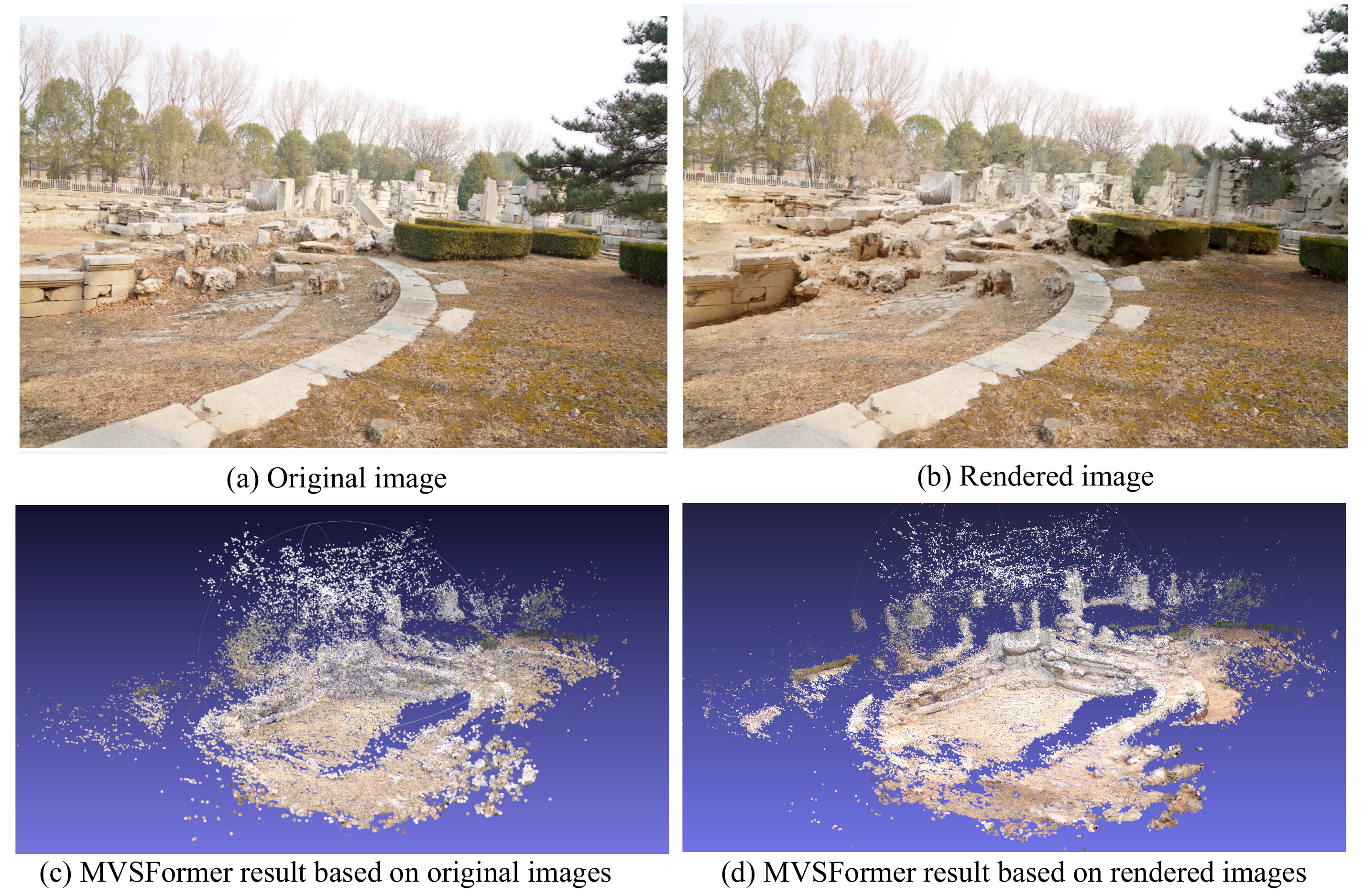} 
  \caption{MVSFormer's results based on original images (c) and rendered images from~Sec.~\ref{sec:render} (d). Note that the MVSFormer is not fine-tuned, and all other settings are unchanged (camera poses, depth ranges, image scale).}
  \label{fig:render_mvs}
\end{figure}

In this section, we present a significant finding from our experiments: \textit{rendered images with Lambertian material result in better reconstruction for learning-based MVS methods, especially in the large scene.} 
We provide MVSFormer results (without any fine-tuning) for both original and rendering fused images in Fig.~\ref{fig:render_mvs}, clearly showing that MVSFormer based on rendered images performs much better even without any fine-tuning. 
Notably, all settings are identical (camera poses, depth ranges, image scale) except the image, indicating that the significant performance gap is caused by the illumination of rendered images.

As mentioned in MVSNet~\cite{yao2018mvsnet}, non-Lambertian materials in the real-world scene have highly diverse and intricate specular and diffuse reflections (Fig.~\ref{fig:render_mvs}(a)), which become more challenging in large-scale MVS problems. However, rendered images get rid of this dilemma because they are rendered with manually set illumination. Besides, their materials are all based on the normal mesh without any extra settings. Thus these rendered images contain illumination consistency, \textit{i.e.}, Lambertian, as shown in Fig.~\ref{fig:render_mvs}(b).

Some studies have recognized this issue~\cite{chang2022rc,yamashita2023nlmvs}, but they have not addressed it completely. Furthermore, our findings demonstrate that the illumination consistency significantly affects the generalization of MVS methods in large-scale real-world scene reconstruction.
Our solution, as presented in this report, is based on instance-level reconstruction augmentation through a traditional MVS pipeline (RealityCapture). 
To tackle this problem more effectively, we propose future work in the form of proper pre-training strategies based on inverse rendering and learning-based data augmentation.

\section{Quantitative Results}

We show some quantitative comparisons of OpenMVS in Tab.~\ref{tab:openmvs_results}, while the ensemble results are in Tab.~\ref{tab:upload_results}.
From Tab.~\ref{tab:upload_results}, the ensemble-based OpenMVS largely improves the recall of MVSFormer\&ACMMP. Moreover, rendering fine-tuned MVSFormer further improves the performance.
Furthermore, we find the rendering-based model enjoys better reconstruction with 4k rendered images. Besides, we also empirically filter outliers with rule-based methods, which improves the precision slightly.
Unfortunately, as the submission track is closed and we do not have the ground truth of testing data, it is not easy to provide more ablation experiments to reveal each component of our model.

\begin{table}
\small 
\caption{Quantitative results of OpenMVS.
\label{tab:openmvs_results}}
\centering
\begin{tabular}{c|ccc} 
\toprule 
OpenMVS(single) & Precision & Recall & F-score
\tabularnewline
\midrule
OpenMVS-2k-fuse3-cpu & \textbf{29.772149} & 8.76127 & 12.47284 \tabularnewline
OpenMVS-2k-fuse2-cpu& 
26.787811 & 15.66706 & 17.78673
\tabularnewline
OpenMVS-4k-fuse2-cpu& 
27.63134 & \textbf{15.99978} & \textbf{17.92869}
\tabularnewline
\bottomrule
\end{tabular}
\end{table}

\begin{table}
\tiny 
\caption{Quantitative results of different ensemble models. Our final submission is in \textcolor{red}{red}, and no CaptureReality's point clouds are included. MVSFormer(R-2k) and (R-4k) indicate the rendering fine-tuned MVSFormer based on 2k and 4k inference respectively.
\label{tab:upload_results}}
\centering
\setlength{\tabcolsep}{1.0mm}{
\begin{tabular}{ccccccccc}
\toprule 
MVSFormer\&ACMMP  & OpenMVS  & MVSFormer(R-2k)  & RealityCapture  & MVSFormer(R-4k) & Filter & Precision  & Recall  & F-score\tabularnewline
\midrule 
\checkmark  &  &  &  &  &  & 26.058198  & 11.04043  & 13.58623\tabularnewline
\checkmark  & \checkmark  &  &  &  &  & 25.585472  & 19.47133  & 20.27946\tabularnewline
\checkmark  & \checkmark  & \checkmark  &  &  &  & 27.351779  & 20.29213  & 21.11949\tabularnewline
\checkmark  & \checkmark  & \checkmark  & \checkmark  &  &  & 26.566373  & \textbf{22.35344}  & 22.33566\tabularnewline
\rowcolor{red!25}
\checkmark  & \checkmark  &  &  & \checkmark  & \checkmark  & \textbf{28.118945} & 22.09812 & \textbf{22.69246}\tabularnewline
\bottomrule
\end{tabular}}
\end{table}





\bibliographystyle{unsrt}  
\bibliography{references}

\end{document}